\documentclass[runningheads]{llncs}
\usepackage{graphicx}
\usepackage{color}
\usepackage{soul}
\usepackage{multirow}
\usepackage{caption} 
\usepackage[most]{tcolorbox}
\usepackage{varioref}
\usepackage{hyperref}
\usepackage{cleveref}
\begin{document}
\title{One model to rule them all: ranking Slovene summarizers}
%
%
\author{Aleš Žagar
\and
Marko Robnik-Šikonja
}
\authorrunning{A. Žagar \& M. Robnik-Šikonja}
%
\institute{University of Ljubljana, Faculty of Computer and Information Science, \\
Večna pot 113, 1000 Ljubljana, Slovenia \\
\email{\{ales.zagar,marko.robnik\}@fri.uni-lj.si}}

%
%
\maketitle              
\begin{abstract}
Text summarization is an essential task in natural language processing, and researchers have developed various approaches over the years, ranging from rule-based systems to neural networks. However, there is no single model or approach that performs well on every type of text. We propose a system that recommends the most suitable summarization model for a given text. The proposed system employs a fully connected neural network that analyzes the input content and predicts which summarizer should score the best in terms of ROUGE score for a given input. The meta-model selects among four different summarization models, developed for the Slovene language, using different properties of the input, in particular its Doc2Vec document representation. The four Slovene summarization models deal with different challenges associated with text summarization in a less-resourced language.  We evaluate the proposed SloMetaSum model performance automatically and parts of it manually. The results show that the system successfully automates the step of manually selecting the best model.

\end{abstract}
%
%
%

\section{Introduction}
Text summarization identifies the essential information in a document or a collection of documents and presents it in a concise and coherent manner. In spite of the long efforts of natural language processing (NLP), text summarization is still a challenging task. With the explosive growth of digital information, summarizing large volumes of text into a shorter, more manageable form is becoming increasingly important. 

There are two main approaches to text summarization: extractive and abstractive. Extractive summarization selects a subset of sentences or phrases from the original text that best represents the content. The selected sentences are combined to form a summary.  Abstractive summarization, on the other hand, generates new sentences that capture the meaning of the original text. Extractive summarization is simpler and faster than abstractive summarization, but it can result in summaries that contain redundant and repetitive content. Abstractive summarization is more challenging and requires more advanced natural language processing techniques, but it can produce human-like summaries.

State-of-the-art technology for text summarization has seen a significant shift in recent years with the rise of transformer neural network architectures, such as T5 \cite{raffel2020exploring} and GPT-3 \cite{brown2020language}. This resulted in the summarization models whose summaries closely resemble those written by humans, with few repetitions and inaccuracies. These models are also capable of processing increasingly long content, enabling the creation of summaries for larger volumes of text. Consequently, state-of-the-art automatic summaries can be clear and easy to comprehend for end-users.

In the context of the less-resourced morphologically-rich Slovene language, text summarization is even more challenging than in English, due to limited availability of resources and data, as well as research. 
We produced four Slovene summarization models with different properties and trained them on different training data.\footnote{Within the scope of the RSDO project: \href{https://www.cjvt.si/rsdo/}{https://www.cjvt.si/rsdo/}}. 
Our four models encompass two extraction summarizers (one based on a simple word frequency sentence selection, the other being graph-based), an abstractive T5-based model, and a hybrid extractive-abstractive model. In general, the T5-based transformer model works best but may not generalize well for all types of input text. Therefore, we address the problem of which summarization model is the most appropriate for a given text, based on text length and genre. 

We propose a novel  Slovene summarization system (named SloMetaSum), consisting of extractive, abstractive, and hybrid summarizers and a meta-model that selects among them. The proposed meta-system consists of a fully connected neural network that analyzes the input content and recommends the most suitable summarization model for a given text. To achieve this, SloMetaSum uses the Doc2Vec \cite{le2014distributed} numerical representation of documents and predicts the ROUGE scores for each of the summarizers. By using a combination of approaches, the system can effectively generate high-quality summaries that are informative and easy to understand for many types of text, regardless of their length and genre.
\footnote{The demo is available at https://slovenscina.eu/en/povzemanje. The code repositories are available at https://github.com/azagsam/metamodel and https://github.com/clarinsi/SloSummarizer.}.

Our contributions are: 
\begin{itemize}
    \item We have developed four summarization models that can effectively summarize text of varying lengths and genres, making them versatile for a range of applications.
    \item We overcame the challenges of the low-resourced Slovene language, and created high-performing models for summarizing Slovene text.
    \item We have also created a meta-model that can recommend the best-suited summarization model for a given text based on factors such as length, complexity, level of abstraction, and intended use case.
\end{itemize}

The rest of the paper is organized as follows. We present related research in Section 2. Section 3 describes the datasets. In Section 4, we describe summarization systems and the meta-model. We present our experiments and discuss the findings in Section 5. Section 6 concludes and recommends future research.

\section{Related work}
Early approaches to text summarization relied on statistical frequencies of words, sentence position, and sentences containing keywords \cite{Nenkova05theimpact}. These approaches aimed to extract important sentences or phrases from a text and generate a summary by concatenating them. Abstractive methods involved deleting less important words from the text to create a summary \cite{knight2002summarization}.

Graph-based methods have been another popular approach to text summarization. In this approach, the document is represented as a graph, where sentences are nodes, and edges represent the relationships between them. The graph is then used to generate a summary by selecting the most important sentences. This method has been explored in several works \cite{mihalcea2004textrank}, \cite{erkan2004lexrank}.

With the advent of neural networks, there has been an increasing interest in developing abstractive summarization techniques. Early neural abstractive systems used methods such as LSTM and other recurrent neural networks \cite{see_get_2017}, \cite{nallapati_abstractive_2016}. However, transformer-based architectures have emerged as state-of-the-art models for abstractive text summarization \cite{zhang2020pegasus}, \cite{lewis2020bart}. These models use self-attention mechanisms to selectively focus on important parts of the text and can generate more fluent and coherent summaries compared to earlier methods.

While several approaches have been proposed for text summarization, many of them are designed to handle specific genres or types of text. In this work, our goal is to build a summarization system that can handle every type of text and genre with every possible property that can appear in the real world. This includes texts of varying lengths, topics, styles, and summaries that capture the most important information in the text. Achieving this goal requires developing a robust and adaptable model that can learn to summarize texts of diverse types and produce high-quality summaries.

\section{Datasets}
In this section, we describe the datasets we used in our research. Below, we provide a short description of the datasets, with their statistics contained in Table \ref{tab:datasets}. 

The STA dataset (general news articles from the Slovenian Press Agency) consists of 366,126 documents and the first paragraph of each article was used as a proxy for summary since the dataset does not contain hand-written human summaries. This is a common technique in text summarization, especially in languages that do not have dedicated news article summarization datasets such as English. 

AutoSentiNews \cite{11356/1109} is a similar dataset to STA, consisting of 256,567 articles from the Slovenian news portals 24ur, Dnevnik, Finance, RTVSlo, and Žurnal24. The summaries are produced from the first paragraph in the same way as they are in the STA dataset. 

The SURS dataset is a small financial news dataset from the Slovenian statistical office and consists of 4,073 documents. 

The KAS corpus of Slovene academic writing \cite{11356/1448} consists of BSc/BA, MSc/MA, and PhD theses written from 2000 - 2018 and gathered from the digital libraries of Slovene higher education institutions via the Slovene Open Science portal \footnote{\href{http://openscience.si/}{http://openscience.si/}}. The corpus contains human-written abstracts of academic texts.  

CNN/Daily Mail dataset \cite{hermann2015teaching} is for text summarization. It has human-generated abstractive summary bullets from news stories on CNN and Daily Mail websites. The corpus has 286,817 training pairs, 13,368 validation pairs, and 11,487 test pairs. The source documents have 766 words and the summaries consist of 53 words on average. We translated the dataset in Slovene using machine translation \cite{11356/1739}.

\begin{table}[]
\centering
\begin{tabular}{l|l}
\textbf{Dataset} & \textbf{Number of documents}   \\ \hline
STA              & 334,696          \\
AutoSentiNews    & 256,567          \\
SURS             & 4,073            \\
KAS              & 82,308           \\ \hline
Total            & \textbf{677,644}
\end{tabular}
\caption{Corpora and datasets used to train a Doc2vec document representation model and the meta-model.}
\label{tab:datasets}
\end{table}

\section{The summarization models and the meta-model}
In this section, we describe the components of our SloMetaSum system which consists of four summarization models, a technique for document representation, and the meta-model. 

\subsection{Summarization models}
We produced four summarization models, described below. 

\textbf{Sumbasic} \cite{Nenkova05theimpact} uses a simple word frequency approach to select the most informative sentences. The \textbf{graph-based} summarization model \cite{zagar-robnik-sikonja-2021-unsupervised} was inspired by the TextRank algorithm \cite{mihalcea2004textrank} and uses centrality scores of sentences to rank them. Both models belong to extractive methods and can be used on documents of any size. In contrast to the original TextRank,  we used the transformer-based LaBSE sentence encoder \cite{feng2022language}, to numerically represent sentences. The \textbf{T5-article} abstractive summarization model uses a pre-trained Slovene T5 model \cite{ulvcar2022sequence} and is fine-tuned on a machine-translated CNN/Daily Mail dataset \cite{hermann2015teaching} using the Slovene machine translation system \cite{11356/1739}. The \textbf{hybrid-long} summarization model is a combination of the graph-based and the T5-article model. It first constructs a short text by concatenating the most informative sentences (extractive step). In the next, abstractive step, these sentences are summarized with the T5-article summarizer. 

\subsection{Doc2Vec model representation}
To select the most suitable summarization method for a given text, the meta-model has to get information about different text properties.
We apply the Doc2Vec model for document representation and train it on the Slovene documents presented in Table \ref{tab:datasets} (without abstracts). In the preprocessing step, we removed high-frequency words that do not contribute to the meaning of a document, such as pronouns, conjunctions, etc.; to further reduce the number of different words, we lemmatized the whole dataset. 

\subsection{Meta-model}
Our meta-model consists of a fully connected neural network, trained to predict the ROUGE scores of the summarizers. For a training dataset, we randomly selected 93,419 examples from the raw concatenated dataset. After that, each of our four summarizers produced a summary for all examples. We calculated ROUGE scores between the reference and generated summaries. ROUGE (Recall-Oriented Understudy for Gisting Evaluation) is a metric most commonly used for the evaluation of automatically generated text summaries. It measures the quality of a summary by the number of overlapping units (n-grams, sequences of texts, etc.) between summaries created by humans and summaries created by summarization systems. ROUGE is not a single metric but a family of metrics. The most commonly used are ROUGE-N and ROUGE-L. The first measures the overlapping of n-grams (typically unigrams and bigrams), while the second measures the longest common subsequence found in both summaries. As an input to our meta-model, we use four ROUGE F1-scores (ROUGE-1, ROUGE-2, ROUGE-L, ROUGE-LSum) that show how good the generated summaries are. We split data into train, validation, and test sets in ratios of 90:5:5. 

The sizes of both datasets are presented in Table \ref{tab:training_datasets}. In Table \ref{tab:dataset_averages}, we present the average ROUGE values of our summarizers on long and short texts. Summarizers that are specialized for short texts achieve better results on short texts and vice versa.

\begin{table}[]
\centering
\begin{tabular}{l|l}
\textbf{Model}     & \textbf{Training size} \\ \hline
Doc2Vec   & 677,644        \\
Meta-model & 93,419        
\end{tabular}
\caption{Number of training samples for each model.}
\label{tab:training_datasets}
\end{table}

\begin{table}[]
\centering
\begin{tabular}{l|l|l|l|l}
      & t5-article     & sumbasic & graph-based    & hybrid-long \\ \hline
Short & \textbf{14,01} & 13,11    & 13,15          & 12,55       \\
Long  & 10,51          & 13,12    & \textbf{17,71} & 17,59      
\end{tabular}
\caption{Summarizers ROUGE scores for long and short texts. The best scores for short and long texts are in bold.}
\label{tab:dataset_averages}
\end{table}

\section{Results}
In this section, we present our results and evaluation. We report the performance of the Doc2Vec model and Meta-model in each separate subsection. 

\subsection{Doc2Vec}
We used the following hyperparameters for training the Doc2Vec document representation model: the maximum allowed vocabulary size is 100,000, the size of the vector used for word representation is 256, the window size of context words is 5, the minimum frequency of a word to be included in the vocabulary is 1, and the total number of epochs or iterations for training the model is 5.

We evaluated the Doc2Vec model using manual and automatic techniques. For manual analysis, we inspected the top 3 most similar returned documents for each of a few randomly chosen samples using the cosine similarity and observe whether the topics of the documents overlap. The topics of the documents were similar in most cases and based on that we concluded that the model works as expected. The automatic evaluation was part of the whole pipeline, where the model hyperparameters were tuned to optimize the loss of the meta-model. 

\subsection{Meta-model}
Our final results are presented in Table \ref{tab:results}. We compared the proposed meta-model selection mechanism with three baselines. The \emph{Mean-baseline} model simply takes the predictions for each summarization model and averages them. The highest-scoring model is always selected. The \emph{Tree} uses a regression tree; using the hyperparameter grid search, the minimum number of samples required to split an internal node is 100. The \emph{Forest} method uses a random forest; we experimented with similar values as for the Tree model and set the number of tree estimators to 300. 

Our best model is a neural network with two hidden layers. The hidden layers contain 1024 neurons, and we used a validation split of 0.1 during the training process. The activation function used for this model is the rectified linear unit (ReLU). In addition, for the early stopping scheduling strategy, we set the patience parameter to 2. The loss function utilized for this model is the mean squared error.

Meta-model stopped learning after 7 epochs and performed almost 15 points above Mean-baseline on the test set. We observed that choosing different hyperparameters does not seem to significantly affect the results. We experimented with different hidden layer sizes, numbers of units, and activation functions. We also tried different max vocabulary and window sizes of the Doc2Vec model. We report only the values of the best model. 

Overall, this model was found to be the most effective among the meta-model selection strategies we tested. The high number of neurons in the hidden layer likely contributed to its superior performance, as it allows for a greater degree of complexity in the model's representation of the data. 

We further experimented with two variations of the meta-model. Meta-model-length adds another input neuron that explicitly encodes the input length. We found that this does not improve the model and hypothesize that academic texts are of different genres and the document embedding technique covers it well already. We also tried to balance data since the original dataset contains a 1:5 ratio of long to short texts which rises a potential issue of overfitting on short texts. We reduced the number of short texts in a training set to get a balanced dataset of 16,932 samples for our Meta-model-balanced model. This resulted in a worse-performing model but still better than Mean-baseline.  

Table \ref{tab:rec_frequency} shows the frequencies of how many times each model was recommended by a meta-model out of 1000 samples from a test set. We can see that the t5-article model was recommended the most, with a count of 595 out of 1000 samples. The hybrid-long model was recommended 254 times, followed by the Sumbasic model, which was recommended 80 times. The graph-based model was recommended the least, with a count of 71 out of 1000 samples.

\begin{table}[h]
\centering
\begin{tabular}{c|c}
Model & Count \\ \hline
t5-article & 595 \\
hybrid-long & 254 \\
sumbasic & 80 \\
graph-based & 71 \\ \hline
Total & 1000 \\
\end{tabular}
\caption{Frequencies of how many times each model was recommended by the meta-model out of 1000 samples from the test set.}
\label{tab:rec_frequency}
\end{table}

\begin{table}[]
\centering
\begin{tabular}{l|ll}
\textbf{Model}  & \textbf{Mean squared error} &  \\ \cline{1-2}
Mean-baseline & 84.493      &  \\
Tree & 81.631 & \\
Random forest &  74.975 & \\ \hline
Meta-model-baseline       & \textbf{70.066 }     & \\
Meta-model-length       & 70.146       & \\
Meta-model-balanced       & 79.044      & \\

\end{tabular}
\caption{Results of our four models on the test set. Meta-model-baseline showed significant improvement over Mean-baseline and tree methods. Encoding the length feature explicitly or balancing the dataset did not improve the results.}
\label{tab:results}
\end{table}

\begin{table}[h]
\centering
\begin{tabular}{l|ccc|c}
\textbf{Method} & \textbf{Precision} & \textbf{Recall} & \textbf{F1-score} & \textbf{Support} \\ \hline
t5-article     & 0.33               & 0.11            & 0.16              & 1069             \\
hybrid-long    & 0.25               & 0.34            & 0.29              & 817              \\
sumbasic       & 0.28               & 0.10            & 0.15              & 1196             \\
graph-based    & 0.38               & 0.67            & 0.48              & 1589             \\ 
\end{tabular}
\caption{Classification report. The table includes precision, recall, and F1-score for each method, as well as the number of instances in the test set (Support). The methods include t5-article, hybrid-long, sumbasic, and graph-based. Test accuracy was 0.34.}
\label{tab:classification_report}
\end{table}

According to Table \ref{tab:classification_report}, the graph-based method achieved the highest F1-score of 0.48, with a precision of 0.38 and recall of 0.67. The hybrid-long method achieved  F1-score of 0.29, with  precision 0.25 and recall 0.34. The sumbasic method produced F1-score of 0.15, precision 0.28, and recall 0.10. Finally, the t5-article method achieved the lowest F1-score of 0.16, with  precision of 0.33 and recall of 0.11. Overall, the test accuracy for all methods combined was 0.34.

\subsection{Meta-model vs. the rest}
In Table \ref{tab:final_scores}, we present the final evaluation results obtained from our experiments on the test set. It is noteworthy that the proposed Meta-model outperformed all other models across all ROUGE scores. This result highlights the effectiveness and superiority of the Meta-model in selecting the most suitable summarization approach for a given text. This outcome showcases the potential of our approach in automating the process of selecting the best summarization model, eliminating the need for manual intervention. 

\begin{table}[h]
\centering
\begin{tabular}{l|lll}
\textbf{Model} & \textbf{ROUGE-1} & \textbf{ROUGE-2} & \textbf{ROUGE-L} \\
\hline
t5-article & 19.01 & 5.61 & 13.52  \\
graph-based & 19.47 & 5.52 & 12.50  \\
hybrid-long & 18.55 & 5.42 & 11.73 \\
sumbasic & 18.86 & 5.04 & 12.25 \\
Meta-model & \textbf{20.38} & \textbf{5.85} & \textbf{13.67} \\
\end{tabular}
\caption{Performance on the test set for all models. Meta-model achieves the best results in all three categories. }
\label{tab:final_scores}
\end{table}

\section{Conclusion}
In this paper, we proposed a novel system for extractive, abstractive, and hybrid summarization tasks. Our system consists of a trained fully connected neural network that analyzes the input content and recommends the most suitable summarization model for a given text. This approach addresses the problem of selecting the appropriate model for a new text, which can be short, long, and of various genres, and can come from almost anywhere when used in production. Our system provides a more effective and efficient way of generating high-quality summaries for Slovene texts. 

While the proposed SloMetaSum model presents an innovative solution to the problem of selecting the most suitable summarization model for a given text, it is not without its weaknesses. One major drawback is the reliance on the ROUGE score as the sole criterion for model selection. While ROUGE is a commonly used metric in the field of text summarization, it does not always accurately reflect the quality of a summary or capture its coherence and readability. Another potential weakness is the limited scope of the study, which focuses exclusively on the Slovene language. While the four summarization models developed for Slovene are an important contribution to the field, they may not generalize well to other less-resourced languages since it requieres a good automatic translation system.

Future work could involve extending this system to other languages. Another area for future work could involve comparing the proposed system with recent large language models. In addition to evaluating the technical performance of the system, it would also be useful to conduct user studies to assess its usefulness and effectiveness in real-world scenarios. For example, researchers could design experiments to evaluate the system's ability to summarize news articles, academic papers, and other types of content that people encounter in their daily lives.

\section*{Acknowledgments}
The work was partially supported by the Slovenian Research Agency (ARRS) core research programme P6-0411, as well as projects J6-2581, J7-3159, and CRP V5-2297.

%
%
%
\bibliographystyle{splncs04}
\bibliography{mybibliography}
%




\end{document}